\documentclass{article}

\usepackage{amsmath,amsfonts,amssymb,times,graphicx,natbib,algorithm,algorithmic,hyperref,array,enumitem,subfiles,todonotes}

\usepackage[accepted]{whi2020}

\icmltitlerunning{XAI Methods for Neural Time Series Classification: A Brief Review}

\begin{document}

\twocolumn[
\icmltitle{XAI Methods for Neural Time Series Classification: A Brief Review}


\icmlsetsymbol{equal}{*}

\begin{icmlauthorlist}
\icmlauthor{Ilija \v{S}imi\'{c}}{vis}
\icmlauthor{Vedran Sabol}{vis}
\icmlauthor{Eduardo Veas}{tu}
\end{icmlauthorlist}

\icmlaffiliation{vis}{Know-Center GmbH, Knowledge Visualization Area, Graz, Austria}
\icmlaffiliation{tu}{University of Technology Graz, Austria}

\icmlcorrespondingauthor{Ilija \v{S}imi\'{c}}{isimic@know-center.at}

\icmlkeywords{Machine Learning, Explainable AI, XAI, Time Series Classification, Deep Learning, Review, Survey, ICML}

\vskip 0.3in
]


\printAffiliationsAndNotice{}

\begin{abstract}
Deep learning models have recently demonstrated remarkable results in a variety of tasks, which is why they are being increasingly applied in high-stake domains, such as industry, medicine, and finance. Considering that automatic predictions in these domains might have a substantial impact on the well-being of a person, as well as considerable financial and legal consequences to an individual or a company, all actions and decisions that result from applying these models have to be accountable.
Given that a substantial amount of data that is collected in high-stake domains are in the form of time series, in this paper we examine the current state of eXplainable AI (XAI) methods with a focus on approaches for opening up deep learning black boxes for the task of time series classification. 
Finally, our contribution also aims at deriving promising directions for future work, to advance XAI for deep learning on time series data.

\end{abstract}

\section{Introduction}
\label{introduction}
Machine learning refers to a set of methods used to establish (learn) numerical relationships between data samples aiming at: a) predicting a sample given a set of samples, b) assign a sample to one of known groups of samples. In this context, deep learning models have recently demonstrated remarkable results in diverse tasks, such as image classification \cite{krizhevsky2012imagenet} or machine translation \cite{bahdanau2014neural}.
Moreover, these models are being increasingly applied for tasks where mispredictions might have serious consequences, ranging from industrial predictive maintenance \cite{huuhtanen2018predictive}, over heartbeat anomaly detection \cite{rajpurkar2017cardiologist} to stock market forecasts \cite{chong2017deep}. 

Considering that automatic predictions made by an AI might have a substantial impact on the well-being of a person, as well as considerable financial and legal consequences to an individual or a company, all actions and decisions that result from applying these models have to be accountable. 
Therefore, a lack of interpretability directly hinders the adoption rate of deep learning models in high-stake domains, regardless of their potential extraordinary accuracy.
These accountability and interpretability aspects become even more pressing in the event that a right for explanations becomes mandatory by law \cite{regulation2016regulation}.  

In this paper we examine the current state of AI-interpretability methods - a field of research colloquially known as eXplainable Artificial Intelligence (XAI). 
In particular, we focus on approaches for opening up deep learning black boxes for the task of time series classification. Given the very promising, recent developments of deep learning models for classifying time series, and considering that a substantial amount of data that is collected in high-stake domains (industry, medicine, finance) are in the form of time series, this appears as both an exciting, as well as challenging research area. 

We first give a short introduction to the main deep learning architectures related to the task of classifying time series, and highlight the most promising variants according to recently published literature. Following that, we briefly explain what interpretability is and motivate why it is needed. As our central contribution, we examine recently published literature, which deals directly with interpretable machine learning and XAI methods regarding neural time series classification. To conclude, we discuss opportunities for further research in this domain, and derive directions for future work.

\section{Deep Learning Architectures for Time Series Classification}
\label{dl_models_for_tsc}
Time series data refers to data that are ordered in time, such that each data point is recorded with its time relationship. Time series data can have one (univariate) or more (multivariate) channels with varying update rates. The task of time series classification seeks to assign a sequence of data points to some type of known event.

Within machine learning, deep learning refers to algorithms that use a cascade of non-linear computational units (neurons) which allows them to learn hierarchical representations of the data. 
However, one caveat of deep learning is that the learned models cannot be understood by inspecting the model's internals. 

\subsection{Common Architectures}
One of the best-known architectures in deep learning is the multi-layer perceptron (\textbf{MLP}) \cite{lippmann1987introduction}. 
An MLP consist of several layers, where every neuron of a layer is connected to every neuron in the subsequent layer. 
Even though MLPs may function as universal function approximators \cite{cybenko1989approximation}, they have difficulties modeling the dependencies between time steps of a time series, since they treat each input feature independently by training the weights for all the features separately. Nevertheless, MLPs still have a purpose in time series classification, since the accuracy that can be achieved with an MLP can be used as a baseline to evaluate more sophisticated deep learning architectures. 

To allow a neural network to model temporal behavior, recurrent neural networks (\textbf{RNN}) were proposed.
An RNN makes sequential predictions for each individual time step, while taking into account the predictions of the previous time steps. Since the weights across the neurons in an RNN are shared, it can generalize better over time, as well as process inputs of variable length without increasing the model size.
Although RNNs were specifically designed to work on data with sequence dependencies, such as text or sound, they also have their shortcomings.
They are relatively slow to train, due to the necessary sequential computation for each individual time step.
Additionally, they have limitations in modelling long sequences, for which variants of RNNs have been proposed, namely \textbf{LSTMs} \cite{hochreiter1997long} and \textbf{GRUs} \cite{cho2014learning}. Furthermore, traditional RNNs do not take into account future time steps, which for time series classification may prove beneficial. To address this issue, bidirectional variants of the previously mentioned networks have been introduced.

An additional deep learning architecture that is suited for time series classification is a convolutional neural network (\textbf{CNN}) \cite{lecun1989backpropagation}. Despite the fact that CNNs were originally proposed for classifying images, they were shown to be well suited for the task of time series classification \cite{fawaz2019deep}.
CNNs work by applying convolutional filters on the whole time series utilizing a sliding window approach. 
Since the trained weights of a filter are shared across all the time steps of the time series, and since the convolutional operation of a time window is independent of any previous computations, the training of CNNs is very fast, due to the possible parallelization.
An additional benefit of CNNs is that they are exceptionally noise-resistant, which makes feature engineering less of an issue. In the case of multivariate time series, they are also able to capture cross channel correlations.

\subsection{Recommendations from Literature}
Wang et al. \cite{wang2017time} set a baseline for what deep learning models should be able to achieve. They compared MLPs, as well as CNN variants, namely fully convolutional networks (\textbf{FCNs}) and residual networks (\textbf{ResNets}) to non-deep learning state-of-the-art models at that time, such as \textbf{COTE} \cite{bagnall2015time} and \textbf{BOSS} \cite{schafer2015boss}. They showed that deep learning models can achieve competitive accuracy's, when compared to the state-of-the-art models of traditional methods.

Fawaz et al. \cite{fawaz2019deep} supported this claim in their extensive study, where they trained 8730 deep learning models on 97 time series datasets, to evaluate their suitability for time series classification, as well as to compare them to other state-of-the-art models. 
The results of the study have clearly shown that ResNets and FCNs were best suited for the time series classification task when compared to MLPs and variants of RNNs. Furthermore, they compared ResNet, the best performing deep learning model in their study, to the best non-deep learning models for time-series classification at the time. Their findings were that ResNets can compete, although slightly less accurate, with the state-of-the-art non-deep learning models.
However, the measured difference in accuracy when comparing the results of ResNet with the best performing non-deep learning model, namely \textbf{HIVE-COTE} \cite{lines2018time}, which is an ensemble of 37 classifiers, was not significant.
More recent work by Fawaz et al. \cite{fawaz2019inceptiontime} introduced \textbf{InceptionTime}, a CNN that utilizes inception modules \cite{szegedy2015going} for time series classification, achieving higher accuracy than ResNets.

Even though deep learning approaches do not yet beat other state-of-the-art models for time-series classification, they do come very close, with the recent work clearly demonstrating substantial potential for improvement. Also, deep learning exhibits certain advantages which are not shared by traditional methods.
One of the most powerful properties of deep learning models is their ability to learn meaningful representations from raw data, which is a necessary property for end-to-end models that do not require manual feature engineering and prior knowledge about the data. Additionally, due to the heavy parallelization capabilities, CNNs can classify time series very fast, and easily beat HIVE-COTE in regards to inference time.  
Therefore, since deep learning approaches have proven themselves to excel for other data types (e.g. image classification or natural language processing), we believe that their rapid recent advances for time-series classification make them worth of further investigation.

\section{The Need for Interpretability}
\label{interpretability}

Considering that in high-stake domains a model's decision may have a substantial impact on the well-being of a person, as well as considerable financial and legal consequences for a company, a single metric during the model development, such as test set accuracy, is not sufficient to rely on that a model will work as expected when it is in production. While metrics can give an excellent summary of the overall performance, it may also be necessary to understand what exact criteria lead a model to make a certain prediction, by interpreting and analyzing its behavior. 

However, what does interpreting a model, or model interpretability in the context of machine learning mean and why is it actually important? Since this term was used rather freely in the literature, Doshi-Velez \& Kim \cite{doshi2017towards} defined interpretability in the context of machine learning as \textit{the ability to explain or to present in understandable terms to a human}. 

According to Selvaraju et al. \cite{selvaraju2017grad}, interpretability has different uses depending on the quality of a model compared to humans. In the case that a model performs worse than a human, interpretability can be used to identify failure modes. If a model performs equally good as a human, interpretability can be used to verify the model's workings and to foster trust that the model behaves as it should. When a model outperforms a human, interpretability can be used to gain new insights and learn from the model. The need for interpretability and its benefits are therefore evident.

Depending on a model's degree of interpretability a distinction between white-box and black-box models can be made, where the former are considered as inherently interpretable or transparent. Black-box models, of which deep neural networks are the most notable ones, are typically very complicated so that by simply observing the input and a model’s internals we can not understand what the contributing aspects of a model’s prediction have been. Therefore, the role of interpretability methods (also referred to as XAI methods) is to present the contributing factors of a model's prediction in human understandable form.

\section{State of XAI for Time Series}
\label{open_up_bb}

The literature on interpreting black-box models for time series is rather scarce when compared to the approaches for images and text. In a recent evaluation of XAI methods, Guidotti et al. \cite{guidotti2018survey} specifically pointed out that there is a lack of literature addressing the problem of explaining models for time series data, even though they analyzed over 60 methods for opening up black boxes. 
This is also supported by Adadi et al. \cite{adadi2018peeking} and Hohman et al. \cite{hohman2018visual} in their surveys on the state of explainable AI methods, where interpretability methods concerning time series have not even been included, whereas methods for images, text and tabular data were presented and discussed.
Nevertheless, approaches to interpret deep learning models in the domain of time series classification are lately getting more traction. 

Wang et al. \cite{wang2016representation} used a \textbf{deconvolutional network} \cite{zeiler2010deconvolutional} to learn meaningful representations in an unsupervised learning manner. 
They were able to show what the deconvolution network learned by discretizing and converting the representations to a Markov Matrix, so that they can be visualized as a graph. When drawing the graph, they used the force directed placement algorithm which resulted in different typical graph layouts of the representations depending on the class.

Using an autoencoder, Gee et al. \cite{gee2019explaining} proposed an approach to learn class representative prototypes by penalizing prototypes that are too similar to each other with respect to the L2 distance in the latent space. 
They visualize the learned representations by first applying Principal Component Analysis (PCA) on the learned representations of the data to reduce the number of dimensions, after which the cosine similarity is calculated between the reduced vectors so that a similarity matrix is produced. Finally, using t-distributed stochastic neighbor embedding (t-SNE, an algorithm for dimensionality reduction suited for visualizing data) the similarity matrix is reduced to three dimensions for visualization purposes.
The prototypes are visualized from the dimensionally reduced latent space by selecting representations from the class specific clusters (e.g. from the center of mass) in the visualized latent space. After projecting a representation of the reduced vector back to the original latent space, the decoder part of the autoencoder can be used to visualize it. The authors argue that prototypes are representative examples that describe influential features in the latent representations, and provide insight into features across the used training samples.

Goodfellow et al. \cite{goodfellow2018towards} trained a CNN to classify univariate ECG data which represents the heart rhythm of a person. They used class activation maps (\textbf{CAMs}) \cite{zhou2016learning} to highlight the most important aspects of a time series that contributed to a decision. 
CAMs visualize the feature importance by projecting the output layer's weights back on to the last convolutional layer's feature maps.
While, to a certain degree, CAMs offer a simple way to identify feature importance, one main disadvantage of using CAMs is that a global average pooling (GAP) layer is mandatory before the output layer. This limits the application of this XAI method, given that CNNs typically have a number of fully connected layers between the convolutional layers and the output layer, with the purpose of increasing the networks accuracy.
Assaf et al. \cite{assaf2019explainable} utilized a CNN for a multivariate time series regression problem, which consisted of giving a forecast of the average power consumption of a photo-voltaic power plant. To highlight the most important features of the input they employed \textbf{Grad-CAM} \cite{selvaraju2017grad}, which is a generalization of CAM where the GAP layer before the output layer is not necessary.

While proposing a new neural network architecture, the multilevel Wavelet Decomposition Network (\textbf{mWDN}), Wang et al. \cite{wang2018multilevel} additionally presented an \textbf{importance analysis} method for mWDNs. An mWDN decomposes a time series in high and low frequency sub-sequences per network layer in an iterative manner. In the first layer, the input time series is split into a high and low frequency part, which are classified by an intermediate classifier. In the next layer, the low frequency sub-sequence of the previous layer is split again into a high and low frequency part, to which the results of the intermediate classifier of the previous layer is added. The result of this addition is then passed to the next intermediate classifier, and the process is repeated for a defined number of layers.
The importance analysis method resembles \textbf{sensitivity analysis} \cite{simonyan2013deep}, which is a well-known model-agnostic XAI method that measures the importance of each feature by perturbing it and observing the change in the classifiers output. 
The method employed by Wang et al. performs the perturbations in the high and low frequency sub-sequences of the time series in various layers, which makes it possible to assign a relevance to each sub-sequence and layer by observing the output. However, this causes the method to become model-specific. 

Hsu et al. \cite{hsu2019multivariate} used occlusion based \textbf{attention mechanisms} and applied them to time series for interpretable early classification of multivariate time series. The method uses a sliding window with which non-overlapping sub-sequences of a time series are extracted and padded with zeros so that they can be used as an input for a classifier.
A model makes a prediction for each individual sample, which results in a probability distribution over the input windows that can be used to assign a relevance to each sub-sequence.


Finally, in a recent work, Arnout et al. \cite{arnout2019towards} compared multiple existing interpretability methods, namely \textbf{saliency maps} \cite{simonyan2013deep}, Layer-wise Relevance Propagation (\textbf{LRP}) \cite{bach2015pixel}, \textbf{DeepLIFT} \cite{shrikumar2017learning}, Local Interpretable Model-Agnostic Explanations \textbf{LIME} \cite{ribeiro2016should} and SHapley Additive exPlanations \textbf{SHAP} \cite{lundberg2017unified}, and applied them to LSTMs, fully convolutional networks, as well as ResNets. They evaluated the explanations by using the perturbation method, which is a method that perturbs the most relevant input features according to an XAI method and observes the confidence drop of the classifier compared to the original prediction.

Considering that Arnout et al. evaluated multiple XAI methods, we will first briefly describe the idea behind each one. The term saliency maps is often used interchangeably with sensitivity analysis, whereas saliency maps are a visualization after the feature importance has been determined, typically using sensitivity analysis. Nonetheless, it can also be used in combination with other XAI methods. LRP is a method that takes the output value of the network for a specific input, and redistributes it throughout the network back to the input space, while taking into consideration the weights of a neuron and the activations computed for the specific input. DeepLIFT has a similar approach to LRP, by propagating the output of the network back to the input space, however, it redistributes the output value relative to the weights of the model and the activation difference of the neurons compared to a reference sample, which has to be specified for each class individually. LIME is a model-agnostic method, meaning, it can be used to explain any model. LIME generates an explanation by training a local interpretable classifier, with respect to a single prediction. The training data that is needed to create such an interpretable classifier is generated by taking a specific input, permuting it, and labeling the permutations by using the model. Finally, the SHAP method calculates the individual feature importance of the input by calculating the importance of each input feature with respect to an average prediction of the model.
In their experiments, the authors identified that perturbing the relevant input features identified using DeepLIFT and LRP resulted in the largest confidence drop for CNNs, while saliency maps and SHAP worked the best for LSTMs. In the case of ResNets, SHAP identified the most relevant features, while LIME performed worst for all tested architectures.

\section{Discussion}
\label{discussion}
After reviewing the literature on XAI methods for time series classification, it is noticeable that recent contributions increasingly employ and evaluate methods that were originally developed for images. 
This is likely due to the fact that many of the existing XAI methods can be easily transferred to time series data, given that the CNNs that are used for images and time series are essentially the same. 
This line of research is particularly relevant because CNN-related deep learning techniques, such as FCNs, ResNets and InceptionTime, have demonstrated great potential performing at least as good as one of the best traditional time series classification methods (HIVE-COTE), while at the same time providing a significantly higher inference throughput. 

In the following, we highlight the advantages and disadvantages of the individual XAI methods employed for neural time series classification. The methods are divided based on their explanation target, namely feature importance, or model and data properties. The feature importance methods are further subdivided into model agnostic and model specific methods, where the former can be used with any black-box model, while the latter are limited to certain models only. 

\subsection{Feature Importance Methods}
\subsubsection{Model Specific Methods}

Due to the algorithmic simplicity, CAMs are a very easy to implement XAI method. However, their main disadvantage is the necessity of the GAP layer before the output layer of the network, which limits their application drastically. The necessity of the GAP layer was solved with the introduction of Grad-CAM, which is a generalized version of CAM. However, since Grad-CAM is a gradient based method, it suffers of gradient saturation and thresholding problems. The gradient saturation problem occurs when activation functions have ranges where they become flat for high and low values (\emph{sigmoid}, \emph{tanh}), which causes the gradient to be close to zero for high and low values. Similarly, the gradient thresholding problem can cause the gradient to become suddenly zero for activation functions which contain a threshold (\emph{ReLU}). These two problems can distort the actual relevance of the features. Additionally, CAM and Grad-CAM give more of an approximate region of importance of input values for the classifier.

The LRP method can generate excellent explanations with sharp transitions between relevant and irrelevant input features if configured properly. The configuration of LRP consists of defining which one of the multiple relevance propagation rules should be applied at particular layers of the network. This represents an important disadvantage, since the rules have a significant impact on the explanation. Also, LRP is often treated as a non-invasive method, however detailed knowledge about a model architecture is necessary to apply it correctly. Moreover, the rules expect that only ReLU activation functions are used throughout the network.
The DeepLIFT method can also create meaningful explanations, as long as the reference sample used for computing the activation differences is selected well, which can prove to be difficult. Also, any activation function can be used in the network when applying DeepLIFT, since the method propagates the relevance by computing the individual activation differences in the neurons.

\subsubsection{Model Agnostic Methods}
Sensitivity analysis is an easy to apply method, given that only the input values of a sample have to be perturbed, and these perturbed samples predicted by the model. 
A disadvantage of sensitivity analysis is the relatively slow computation time, since the model has to make a prediction after every individual input feature has been perturbed. Additionally, the applied perturbation method can have a significant impact on the output of the model, and the generated heatmaps are often very noisy. Since Wang et al. \cite{wang2018multilevel} applied a model specific variant of sensitivity analysis as an explanation method for their mWDNs, similar drawbacks are to be expected. However, a benefit of their method is that in addition to identifying the importance of individual input features, it is also possible to identify the importance of the individual layers. 

The occlusion-based attention mechanism of Hsu et al. \cite{hsu2019multivariate} is a XAI method where the main advantage is the option to define the sub-sequence sizes, for which the relevance should be computed, instead for individual time steps. 
The main disadvantage is that it might become potentially slow in the case that the relevance for very small sub-sequences would have to be computed. 

Another model agnostic method that was applied is LIME, which creates an explanation by providing an interpretable model that explains a local prediction, which can be a powerful explanation method.  However, the selection of a suitable interpretable surrogate model, the generation of data local to the sample of interest, as well as the training of the surrogate model, might be time consuming and difficult when compared to other methods. 

SHAP is the last model agnostic method analysed by \cite{arnout2019towards}. Since it is not clearly specified by the authors which SHAP variant was employed, the exact benefits and disadvantages cannot be specified. Nonetheless, the feature relevance computed with SHAP is generally fairly distributed across the input features, while the main disadvantage is the relatively slow computation time.

\subsection{Model and Data Properties Methods}
While the feature importance methods focus on probing the model, either by modifying the input and observing the change in the output, or by redistributing the models output back to the input space, autoencoder-based approaches follow a different idea. Autoencoders have been applied for creating representations of raw time series, to analyze how a deep learning model might learn such representations by itself. Gee et al.\cite{gee2019explaining} used them in combination with dimensionality reduction methods (e.g. PCA, t-SNE) to visualize features across the training samples in the latent space. 
Using this method, they could create class representative prototypes to understand what features are representative for a class, and to analyze border cases to improve class separation. 
Wang et al. \cite{wang2016representation} used representations learned by a deconvolutional network, which they converted to a graph to show that there is a significant difference in the statistical properties of the graphs between different classes. Nevertheless, a description of how this method can be used to gain information about the data was not given. 

\subsection{Final Remarks}
In general, model specific methods are faster, since they propagate the relevance from the output back to the input through the network in a single pass. Model agnostic methods on the other hand can be used with any black box, however, they have to modify the input in various ways and make predictions on the changed inputs to determine the importance of the features, which makes them slow. 
Regarding the visual representation of feature importance, most of the methods use heatmaps, while some employ line charts with highlighted sub-sequences, or an independent line chart below the time series showing the relevance of each time step.

To our best knowledge, comparisons of XAI methods for time series appear limited, where a single contribution \cite{arnout2019towards} evaluates and compares saliency maps, LRPs, DeepLIFT, LIME and SHAP methods.
According to the authors, the largest confidence drop for CNNs was observed, when the perturbed input features were identified as relevant by LRP and DeepLIFT. For ResNets, using SHAP lead to the fastest confidence drop, which was also the case for LSTMs, for which also saliency maps worked very well. However, according to \cite{fawaz2019deep} LSTMs did not perform as well as CNN-based approaches, so that their application might be limited, e.g. to early classification. 
LIME performed worst for all the tested architectures, with the possible reason being the difficulty of creating meaningful data that is necessary to train the local interpretable classifier. Finally, many of these methods have been evaluated with the same publicly available datasets, such as the UCR archive \cite{UCRArchive}, UEA archive \cite{bagnall2017great} and Baydogan’s archive \cite{BaydogansArchive}, which supports method comparability. However, evaluations with real-world data and real users are neglected although XAI methods should, ultimately, address humans.

\section{Future Work}
\label{future_work}
The research on interpretability methods for black-box models for time series classification has been gaining traction recently. Therefore, it is currently an area with exciting research opportunities that wait to be explored. Considering that there are types of data for which interpretability methods have been explored much more thoroughly (e.g. images or text), transferring and adapting these methods, and applying them to the task of time series classification appears as a promising starting point. 

In addition to the development of methods for identifying feature importance, the way of communicating these relevances ought to be explored in more detail. For example, given an image, a heatmap might be a suitable representation of feature importance, however for long time series with numerous channels this kind of visual representation might not be suitable. Thus, other representation forms specific to time series data sets and features pertaining to this data type should be explored, including methods which aggregate and summarize time series information or textual explanations of the key contributing factors.

As XAI techniques should ultimately address users, we strongly believe that their performance cannot be expressed solely by metrics computed automatically by probing the model. In addition to the development of new time series-specific XAI methods and adaptations of existing methods for other data types, studies involving real users from high-stake domains are, in our opinion, crucial. User studies must be conceived and performed to assess human-oriented properties of XAI methods, such as the readability and usability of the visual representations, subjective usefulness of the technique, as well as properties contributing to acceptance of deep-learning in practice, like trust and perceived transparency. Therefore, XAI methods should not be only evaluated with synthetic benchmarks, but also in real-life scenarios, with real data and, very importantly, involving the specific target audiences.

We propose to approach XAI for time series data in a user-driven manner. In a first phase, 
an understanding of the working processes and environments, where deep learning already is or could be applied on time series data, shall be obtained through interviews with the experts (stakeholders) from the aforementioned high-stake domains. Moreover, the best suited formats to communicate the information, as well as the expected methods of interacting with a model should be identified.
By the end of this phase, we expect to obtain the answers to the following four questions: i) What expectations do users have of an explanation method? ii) What is the most appropriate format to communicate this information? iii) In what manner can users interact with the explanations? iv) How can explanation methods be integrated into the existing working processes? In a second phase, existing XAI methods will be adapted and new methods developed that address the user needs identified in the first phase. Subsequently, user studies shall be performed to evaluate the employed XAI methods and to learn which further improvements would be most valuable.

\section{Acknowledgement}
\label{Acknowledgement}
This work was supported by the "DDAI" COMET Module within the COMET – Competence Centers for Excellent Technologies Programme, funded by the Austrian Federal Ministry for Transport, Innovation and Technology (bmvit), the Austrian Federal Ministry for Digital and Economic Affairs (bmdw), the Austrian Research Promotion Agency (FFG), the province of Styria (SFG) and partners from industry and academia. The COMET Programme is managed by FFG.

\bibliography{bibliography}
\bibliographystyle{icml2020}

\end{document}